\documentclass[final]{cvpr}

\usepackage{times}
\usepackage{epsfig}
\usepackage{graphicx}
\usepackage{amsmath}
\usepackage{amssymb}
\usepackage{bbding}
\usepackage{float} 
\usepackage{ulem}
\usepackage{booktabs}
\usepackage{colortbl}
\usepackage{xcolor}
\usepackage{subfigure}
\usepackage{lipsum}

\definecolor{mygray}{gray}{.85}

\usepackage[pagebackref=true,breaklinks=true,colorlinks,bookmarks=false]{hyperref}

\def\cvprPaperID{261} 
\def\confYear{CVPR 2023}
\definecolor{myblue}{RGB}{0,176,240}
\definecolor{myred}{RGB}{192,80,70}
\definecolor{purple}{RGB}{112,48,160}
\begin{document}

\title{Disentangling Orthogonal Planes for Indoor Panoramic Room Layout Estimation with Cross-Scale Distortion Awareness}

\author{Zhijie Shen, 
Zishuo Zheng,
Chunyu Lin$^{\dagger}$, 
Lang Nie,
Kang Liao,
Shuai Zheng, and
Yao Zhao\\
Institute of Information Science, Beijing Jiaotong University, China\\
Beijing Key Laboratory of Advanced Information Science and Network Technology\\
{\tt\small zhjshen@bjtu.edu.cn \tt\small cylin@bjtu.edu.cn}
}

\maketitle    
\renewcommand{\thefootnote}{\fnsymbol{footnote}}
\footnotetext[2]{Corresponding author}
\renewcommand{\thefootnote}{\arabic{footnote}}

\begin{abstract}
   Based on the Manhattan World assumption, most existing indoor layout estimation schemes focus on recovering layouts from vertically compressed 1D sequences. However, the compression procedure confuses the semantics of different planes, yielding inferior performance with ambiguous interpretability. 

   To address this issue, we propose to disentangle this 1D representation by pre-segmenting orthogonal (vertical and horizontal) planes from a complex scene, explicitly capturing the geometric cues for indoor layout estimation. 
   Considering the symmetry between the floor boundary and ceiling boundary, we also design a soft-flipping fusion strategy to assist the pre-segmentation. Besides, we present a feature assembling mechanism to effectively integrate shallow and deep features with distortion distribution awareness.
   To compensate for the potential errors in pre-segmentation, we further leverage triple attention to reconstruct the disentangled sequences for better performance. Experiments on four popular benchmarks demonstrate our superiority over existing SoTA solutions, especially on the 3DIoU metric. The code is available at \url{https://github.com/zhijieshen-bjtu/DOPNet}.
   
\end{abstract}
\begin{figure}[!t]
  \centering
  \includegraphics[width=.5\textwidth]{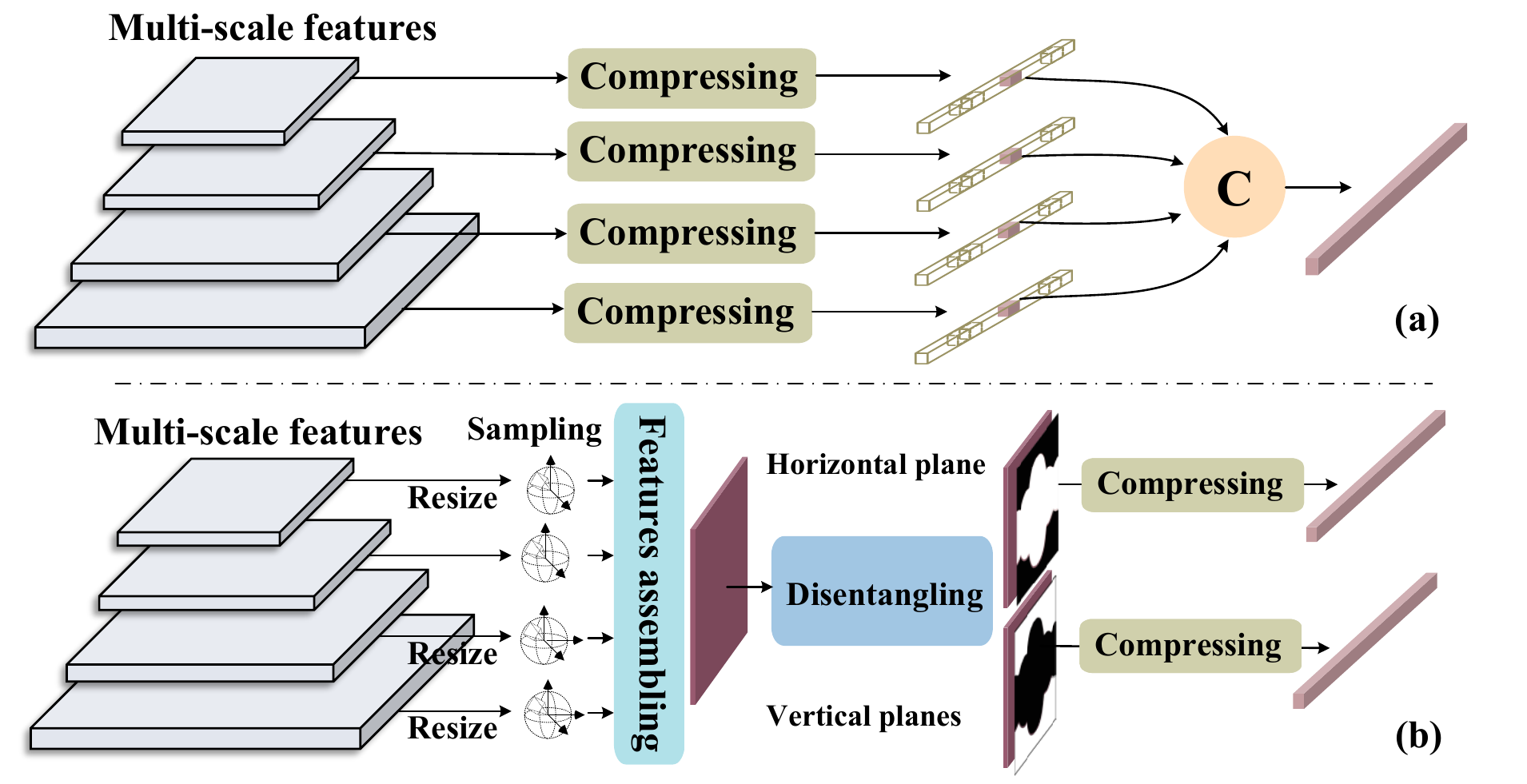} 
  \caption{(a) The commonly used architecture. (b) The proposed one. 
  Compared with the traditional pipeline, ours has two advantages: (1) Disentangling the 1D representation into two separate sequences with different plane semantics. (2) Adaptively integrating shallow and deep features with distortion awareness via a feature assembling mechanism rather than simple concatenation.
  } 
  \label{fig:newarchi}
  \vspace{-0.2cm}
\end{figure}
\section{Introduction}
Indoor panoramic layout estimation refers to reconstructing 3D room layouts from omnidirectional 
images. Since the panoramic vision system captures the whole-room contextual information, we can estimate the complete room layout with a single panoramic image. However, inferring the 3D information from a 2D image is an ill-posed problem. Besides, the 360° field-of-view (FoV) of panoramas brings severe distortions that increase along the latitude. Both issues are challenging for indoor layout estimation.

Different from outdoor scenarios, the indoor room has the following properties: (1) The indoor scenes conform to the Manhattan World assumption (The floors and ceilings are all flat planes, and the walls are perpendicular to them). (2) The room layout is always described as the corners or the floor boundary and ceiling boundary. These characteristics could be used as potential priors to guide 
the design of a reasonable layout estimation method.

Based on the Manhattan World assumption, previous approaches \cite{sun2019horizonnet,sun2021hohonet,Jiang_2022_CVPR,pintore2020atlantanet,wang2021led} prefer to estimate the layout from a 1D sequence. They advocate compressing the extracted 2D feature maps in height dimension to obtain a 1D sequence, of which every element in this sequence share the same distortion magnitude (Fig. \ref{fig:newarchi}{\color{red}a}). 
We argue that this compressed representation does not eliminate the panoramic distortions because there is no explicit distortion processing when extracting 2D feature maps before compression.
Moreover, this strategy roughly mixes the vertical and horizontal planes together, confusing the semantics of different planes that are crucial for layout estimation. 

On the other hand, some researchers devoted themselves to adopting different projection formats to boost the performance, e.g., the bird's view of the room \cite{zou2021manhattan} and cubemap projection \cite{zhao20223d}.
These projection-based schemes weaken the negative effect of the distortions. Nevertheless, frequent projection operations between different formats raise computational complexity. In addition, there exists an inevitable domain gap between the feature maps from different formats.


To address the above problems, we propose a novel architecture (Fig.\ref{fig:newarchi}{\color{red}b}) that disentangles the orthogonal planes in advance to capture an explicit geometric cue. 
Following \cite{wang2021led}, our room layout can be recovered from the predicted horizon-depth map and the room height. Therefore, the ``clean" depth-relevant features and height-relevant features can both help with the layout estimation. 
To obtain such ``clean" features free from the disturbance of decorations and furniture, we disentangle the widely used 1D representation into two separate sequences — the horizontal and vertical ones. Specifically, we pre-segment the vertical plane (i.e., walls) and horizontal planes (i.e., floors and ceilings) from the whole-room contextual information. Then these orthogonal planes are compressed into two 1D representations. Especially, based on the symmetry property between the floor boundary and ceiling boundary, we design a soft-flipping fusion strategy to assist this process. 

Moreover, we propose an assembling mechanism to fuse multi-scale features with distortion awareness effectively. To eliminate the negative effect of distortion, we compute the attention among distortion-relevant positions following distortion distribution patterns.
Meanwhile, cross-scale interaction is carried out to integrate shallow geometric structures and deep semantic features. Considering the error of pre-segmentation, we further leverage triple attention to reconstruct the two 1D sequences. Particularly, we adopt graph-based attention to generate discriminative channels, self-attention to rebuild long-range dependencies, and cross-attention to provide the missing information for different sequences. 


To demonstrate our effectiveness, we conduct extensive experiments on four popular datasets --- MatterportLayout \cite{zou2021manhattan}, Zind \cite{cruz2021zillow}, Stanford 2D-3D \cite{armeni2017joint}, and PanoContext \cite{zhang2014panocontext}. The qualitative and quantitative results show that the proposed solution outperforms other SoTA methods. Our contributions can be summarized as follows:
\begin{itemize}
    \item We propose to disentangle orthogonal planes to capture an explicit geometric cue for indoor 360° room layout estimation, with a soft-flipping fusion strategy to assist this procedure.
    \item We design a cross-scale distortion-aware assembling mechanism to perceive distortion distribution as well as integrate shallow geometric structures and deep semantic features.
    \item On popular benchmarks, our solution outperforms other SoTA schemes, especially on the metric of intersection over the union of 3D room layouts.
\end{itemize}
 
\section{Related Work}
\subsection{360° Room Layout Estimation}
\label{sec2_1}
Based on Manhattan World assumption \cite{coughlan1999manhattan} that all walls and floors are aligned with global coordinate system axes and are perpendicular to each other, many approaches utilize convolutional neural networks (CNNs) to extract useful features to improve layout estimation accuracy. Specifically, Zou $et$ $al.$ \cite{zou2018layoutnet} propose LayoutNet to predict the corner/boundary probability maps directly from the panoramas and then optimize the layout parameter to generate the final predicted results.
Furthermore, they annotate the Stanford 2D-3D dataset \cite{armeni2017joint} with awesome layouts for training and evaluation.
Yang $et$ $al.$ \cite{yang2019dula} propose Dula-Net to predict a 2D floor plane semantic mask from both the equirectangular view and the perspective view of the ceilings.
The modified version, LayoutNet v2, and Du  la-Net v2, which have better performance than the original version, are presented by Zou $et$ $al.$ \cite{zou2021manhattan}.
Fernandez $et$ $al.$ \cite{fernandez2020corners} present to utilize equirectangular convolutions (EquiConvs) to generate corner/edge probability maps.
Sun $et$ $al.$ propose HorizonNet \cite{sun2019horizonnet} and HoHoNet \cite{sun2021hohonet} to simplify the layout estimation processes by representing the room layout with a 1D representation.
Besides, they use Bi-LSTM and multi-head self-attention to build long-range dependencies and refine the 1D sequences.

Recently, several approaches have adopted this 1D representation and achieved impressive performance. For example, Rao $et$ $al.$ \cite{rao2021omnilayout} establish their network based on HorizonNet \cite{sun2019horizonnet}. They replace standard convolution operation with spherical convolution to reduce distortions and take Bi-GRU to reduce computational complexity.
Wang $et$ $al.$ \cite{wang2021led} combine the geometric cues across the layout and propose LED2-Net that reformulates the room layout estimation as predicting the depth of walls in the horizontal direction.
Not constrained to Manhattan scenes, Pintore $et$ $al.$ \cite{pintore2020atlantanet} introduce AtlantaNet to predict the room layout by combining two projections of the floor and ceiling planes.
Driven by the self-attention mechanism, many transformer-based methods \cite{yuan2021tokens, wang2021pyramid, liu2021swin} proposed to build long-range dependencies. For example, Jiang $et$ $al.$ \cite{Jiang_2022_CVPR} represent the room layout by horizon depth and room height and introduce a Transformer to enforce the network to learn the geometric relationships. However, the popular 1D representations could confuse the semantics of different planes, thus causing the layout estimation challenging. We disentangle this 1D representation by pre-segmenting orthogonal planes.

\subsection{360° Image Distortion Mitigation}
\label{sec2_2}
When converting spherical data to the equirectangular projection format, severe distortions are introduced. 
Some recent studies focus on designing spherical customized convolutions to make the network adapt to panoramic distortions and offer superior results compared to standard convolutional networks.
Su $et$ $al.$ \cite{su2017learning} introduce SphConv to remove distortions by changing the regular planar convolutions kernel size towards each row of the 360° images.
Cohen $et$ $al.$ \cite{cohen2018spherical} propose a spherical CNN that transforms the domain space from Euclidean S2 space to a SO(3) group to reduce the distortion and encodes full rotation invariance in their network. 
Coors $et$ $al.$ \cite{coors2018spherenet} addresses it by defining the convolution kernels on the tangent plane and keeping the convolution domain consistent in each kernel. Therefore, the non-distorted features can be extracted directly on 360° images.
Rao $et$ $al.$ \cite{rao2021omnilayout} first applied spherical convolution operation to room layout task and showed significant improvements. Different from the above approaches, Shen $et$ $al.$ \cite{shen2022panoformer} propose the first panorama Transformer (PanoFormer) and divide the patches on the tangent plane to remove the negative effect of distortions. But most recent layout estimation works do not treat the distortions seriously because they think that the 1D representations can solve the distortions well. On the contrary, we propose a feature assembling mechanism with cross-scale distortion awareness to deal with distortions. Extensive experimental results demonstrate the necessity of handling distortions before generating 1D representations.

\begin{figure*}[t]
  \centering
  \includegraphics[width=\textwidth]{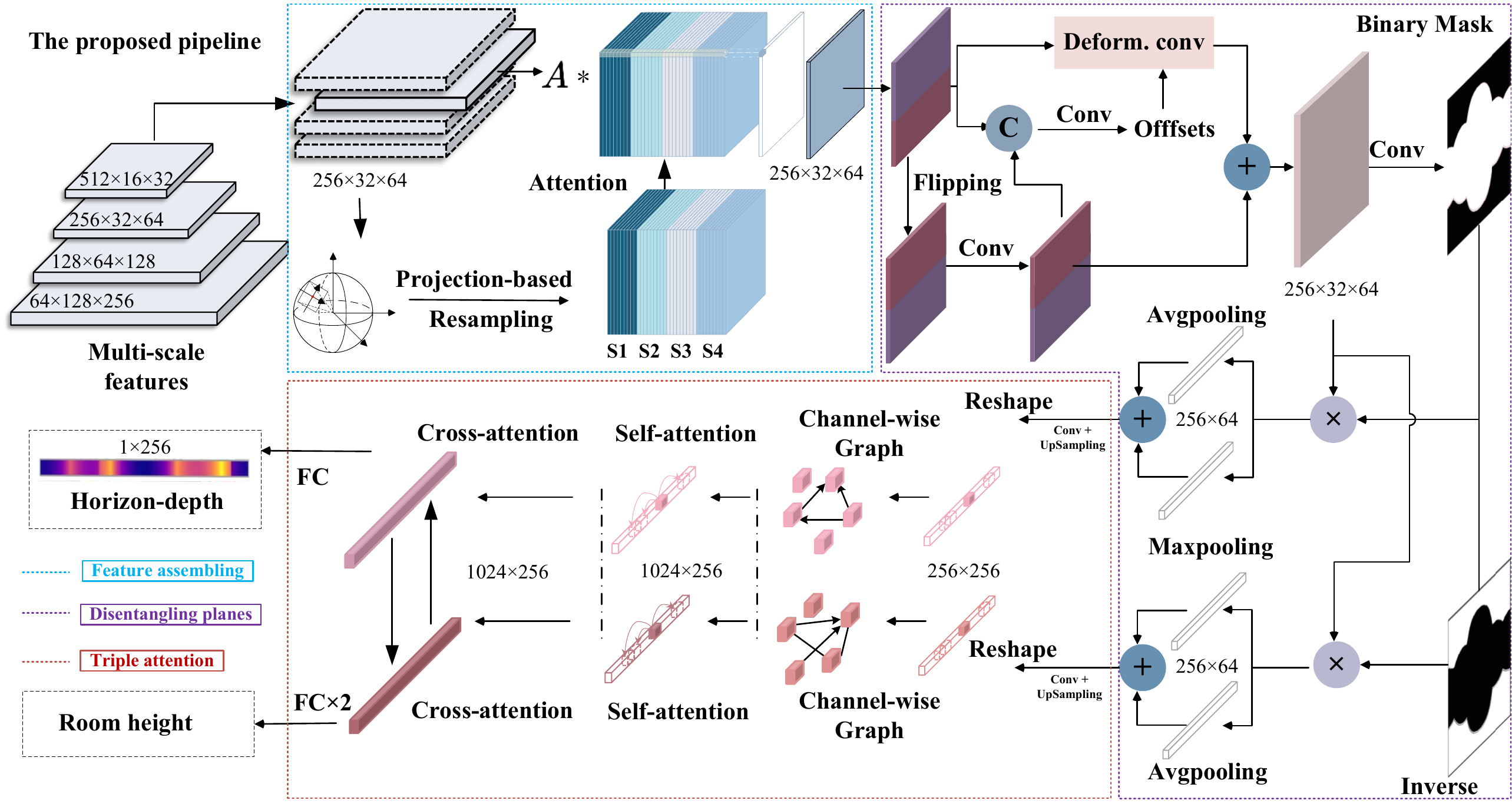} 
  \caption{Overview of the proposed framework. The feature assembling mechanism is designed to deal with distortions as well as integrate shallow and deep features (circled in \textcolor{myblue}{Blue}). Then we disentangle orthogonal planes to produce two 1D representations with distinguished plane semantics (circled in \textcolor{purple}{Purple}). 
  And triple attention is deployed to reconstruct the 1D representations (circled in \textcolor{myred}{Red}). Our scheme leverages a single panorama with a resolution of 512×1024 as input and predicts the room height and the horizon depth.} 
  \label{fig:framework}
\vspace{-0.2cm}
\end{figure*}
\section{Approach}
\label{section3}

\subsection{Overview}
\label{sec3_1}
As illustrated in Fig. \ref{fig:framework}, the proposed framework consists of three stages: feature extraction followed by a feature assembling mechanism, orthogonal plane disentanglement, and 1D representation reconstruction.
Specifically, we adopt a backbone (basically ResNet) to extract a series of features at 4 different scales from a single panorama. Then a feature assembling mechanism is designed to fuse the multi-scale features and free them from panoramic distortions. In the next stage, we introduce a soft-flipping fusion strategy to explore the symmetry between the floor boundary and the ceiling boundary. After that, we segment the vertical and horizontal planes from the fused features followed by vertical compression, yielding two 1D representations. In the last stage, the disentangled features are reconstructed with triple attention to be more discriminative and informative.
Finally, the horizon-depth map and room height are estimated from the reconstructed sequences.

\subsection{Cross-scale Distortion Awareness}
\label{sec3_2}
In previous works \cite{sun2019horizonnet,sun2021hohonet,wang2021led,Jiang_2022_CVPR}, researchers prefer to follow the architecture of HorizonNet \cite{sun2019horizonnet} to extract multi-scale feature maps of an equirectangular image. These feature maps are merged together via vertical compression and concatenation \cite{lin2017feature}. However, the popular architecture just concatenates multi-scale features together and neglects the inherent distortion distribution, resulting in an inferior panoramic feature extraction capability. 
To address the above issues, we propose a novel feature assembling mechanism to deal with distortions, as well as ingeniously integrate shallow and deep features.

\noindent \textbf{Eliminating Distortions.} 
Motivated by previous works \cite{fernandez2020corners,shen2022panoformer}, we first polymerize the most relevant features via a virtue tangent plane on the panoramic sphere projection.
Based on the projection formula \cite{shen2022panoformer,shen2021distortion,Coors_2018_ECCV} (we discuss it in detail in the supplementary material), we can get the distortion-free sampling coordinates $p \in \mathbb{R}^{H\times W\times 9 \times 2}$ (height, width, number of points, number of coordinates) in the 2D feature maps. The learnable offsets $\Delta p$ are employed to adjust the sampling locations adaptively. Then we obtain gathered features $f_{df} \in \mathbb{R}^{C\times H \times W\times 9}$ from the original features $f\in \mathbb{R}^{C\times H \times W}$ as follows:
\begin{eqnarray}
\label{eq1}
\begin{array}{cc}
f_{df}=\sum\limits_{q=1}^{H\times W}\sum\limits_{k=1}^{9}Sample(f, p_{q,k}+\Delta p_{q,k}),
\end{array}
\end{eqnarray}
where $q$ represents the $q^{th}$ point to gather the relevant features; $k$ indexes the $k^{th}$ related features.

\noindent\textbf{Integrating Shallow and Deep features.} We select the $3^{th}$ scale ($\tilde{s}$) as the reference scale and resize other feature maps to this scale. Specifically, we utilize the learnable sampling coordinates ($p_{q,k}+\Delta p_{q,k}$) to gather the distortion-free feature maps for every feature with resized scales following Eq.\ref{eq1}. Let $f^{s}\in \mathbb{R}^{C\times H_{s} \times W_{s}}$ be the original feature map at a certain scale $s$. The gathered multi-scale feature maps $f_{m\tilde{s}} \in \mathbb{R}^{C\times H_{\tilde{s}} \times W_{\tilde{s}} \times 9 \times 4}$ can be represented as:
\begin{eqnarray}
\begin{array}{cc}
&f_{m\tilde{s}}=\sum\limits_{s=1}^{4}\sum\limits_{q=1}^{H_{\tilde{s}}\times W_{\tilde{s}}}\sum\limits_{k=1}^{9} \\&Sample(resize(f^{s},\tilde{s}) , p_{q,k}^{\tilde{s}}+\Delta p_{q,k}^{\tilde{s}}).
\end{array}
\end{eqnarray}
Then, the designed cross-scale distortion-aware feature assembling mechanism can be applied as:
\begin{eqnarray}
\begin{array}{cc}
f_{m\tilde{s}}^{'}=&\operatorname{CSDA}(f_{m\tilde{s}}) =
\sum\limits_{l=1}^{L}A_{l} \cdot reshape(f_{m\tilde{s}}),
\end{array}
\end{eqnarray}
where $L$ indexes the heads of the self-attention; $A_{l} \in \mathbb{R}^{L \times H_{\tilde{s}}W_{\tilde{s}} \times 36}$ is the learnable self-attention weights from $f_{m\tilde{s}}$. To complete the calculation, we reshape $f_{m\tilde{s}}$ into the size of $\mathbb{R}^{L\times H_{\tilde{s}}W_{\tilde{s}}\times 36\times \frac{C}{L}}$. The final assembling features are denoted as $f_{m\tilde{s}}^{'} \in \mathbb{R}^{C\times H_{\tilde{s}}\times W_{\tilde{s}}}$.
\begin{figure}[H]
  \centering
  \includegraphics[width=0.45\textwidth]{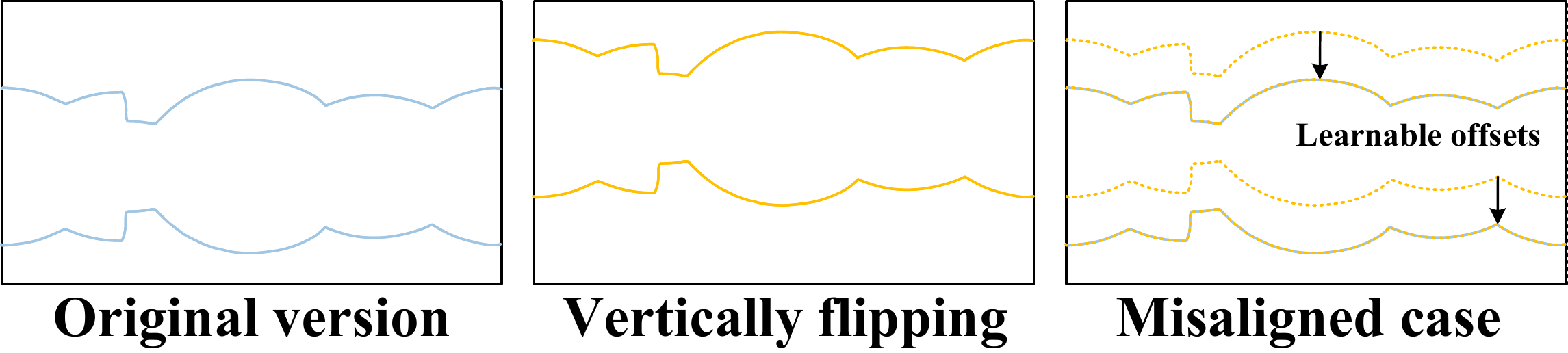} 
  \caption{Illustration of the misaligned case when leveraging the symmetry property.} 
  \label{fig:soft}
\vspace{-0.2cm}
\end{figure}
\subsection{Disentangling Orthogonal Planes}
\label{sec3_3}
In order to explicitly capture the geometric cues, we propose to disentangle the orthogonal planes. Particularly, we segment the vertical plane and horizontal planes from the whole scenario before compressing the 2D feature maps.\\
\noindent\textbf{Symmetry in Manhattan World.}
It is challenging to learn accurate layout boundaries in a complex scene due to the disturbance of furniture and illumination, e.g., occlusion. To address it, we propose to leverage the symmetry of the Manhattan World assumption (e.g., the floor boundary and ceiling boundary are strictly symmetric in 3D space) to provide complementary information.
To exploit this symmetry property, we introduce a soft-flipping fusion strategy.

We first vertically flip the feature maps to get their symmetrical version. In fact, the floor boundary and ceiling boundary are not strictly symmetric in an image because the shooting position is not in the exact middle of the floor and ceiling. To this end, we adopt a deformable convolution with 3×3 kernel size to adaptively adjust the symmetrical version. Next, we fuse the original feature and its ``soft" symmetrical version to provide more informative boundary cues. We denote the fused features as $f_u\in \mathbb{R}^{C\times H_{\tilde{s}}\times W_{\tilde{s}}}$.

 
\noindent\textbf{Segmenting Horizontal/Vertical Planes.} 
To disentangle the 1D representation, we pre-segment the horizon/vertical planes in advance to capture an explicit geometric cue.
Specifically, we generate the pseudo labels for the orthogonal planes with layout GT. Then, we use the generated labels to encourage the network to learn a binary mask 
$S \in \mathbb{R}^{H_{\tilde{s}} \times W_{\tilde{s}}}$ via a simple segmentation head. The two segmented feature maps can be denoted as:
\begin{eqnarray}
\setlength{\abovedisplayskip}{-5pt}
\setlength{\belowdisplayskip}{-5pt}
\begin{array}{cc}  
f_{h}^{p} =& (f_{u} \oplus  f_{m\tilde{s}}^{'})\otimes sigmoid(S),\\
f_{v}^{p} =& (f_{u} \oplus  f_{m\tilde{s}}^{'})\otimes (1-sigmoid(S))
\end{array}
\end{eqnarray}
where $f_{h}^{p}$/$f_{v}^{p}$ represents the horizon/vertical plane features, respectively; $\oplus$ is the element-wise sum and $\otimes$ is the element-wise product. Following previous works, we further vertically compress 2D feature maps to two 1D sequences $\mathbf{Q}_{h}$, $\mathbf{Q}_{v} \in \mathbb{R}^{W_{\tilde{s}} \times C}$ with differnet plane semantics.

\subsection{Reconstructing 1D Representations}
\label{sec3_4}
\noindent \textbf{Generating Discriminative Channels.}
There are generally unconfined dependencies among channels, resulting in confused semantic cues~\cite{graph_reasoning}. To this end, we propose a discriminative channels generation mechanism via graph convolution to enforce each channel to concentrate on distinguishing features. Different from normal pixel-wise~\cite{pixel_level} or object-wise~\cite{object_level} graphs, the introduced channel-wise one tends to guide the node to subtract the information from the neighbor nodes rather than aggregation. The formula of the discriminative channels generation mechanism can be represented as:
\begin{eqnarray}
\begin{array}{cc}  
f' = \mathbf{L}fW = (\mathbf{I}-\mathbf{A})fW
\end{array}
\label{eq::LXW}
\end{eqnarray}
where $\mathbf{L}$ and $\mathbf{A}$ are the symmetric normalized Laplacian matrix and normalized adjacency matrix of the channel-wise graph, and $I$ and $W$ are the identity matrix and learnable weights, respectively. 
Please refer to the supplementary materials for more details about the channel-wise graph attention.
Following Eq. \ref{eq::LXW}, $\mathbf{Q}_{h}$ and $\mathbf{Q}_{v}$ are first reconstructed to be more discriminative in channels (represented as $\mathbf{Q}_{h}'$, $\mathbf{Q}_{v}' \in \mathbb{R}^{W_{\tilde{s}} \times C}$). 

\noindent \textbf{Rebuilding Long-Range Dependencies.} We employ the standard self-attention to reconstruct the intra-sequence long-range dependencies. The attention formula of the horizontal plane sequence can be written as:
\begin{eqnarray}
\begin{array}{cc}   
Attention(\mathcal{Q}_{h}', \mathcal{K}_{h}', \mathcal{V}_{h}') = softmax(\frac{\mathcal{Q}_{h}'(\mathcal{K}_{h}')^{T}}{\sqrt{d_{k}}}\mathcal{V}_{h}'),
\end{array}
\end{eqnarray}
where $\mathcal{Q}_{h}', \mathcal{K}_{h}', \mathcal{V}_{h}'$ are all learned from $\mathbf{Q}_{h}'$ via a fully connected layer.
Similarly, the calculation for the vertical plane sequence is defined as: 
\begin{eqnarray}
\begin{array}{cc}   
Attention(\mathcal{Q}_{v}', \mathcal{K}_{v}', \mathcal{V}_{v}') = softmax(\frac{\mathcal{Q}_{v}'(\mathcal{K}_{v}')^{T}}{\sqrt{d_{k}}}\mathcal{V}_{v}').
\end{array}
\end{eqnarray}
It captures global interactions to adapt to the large FoV of panoramas, contributing to better performance. We denote the two sequences with reconstructed long-range dependencies as $\mathbf{Q}_{h}''$, $\mathbf{Q}_{v}'' \in \mathbb{R}^{W_{\tilde{s}} \times C}$.

\noindent \textbf{Providing Missing Residuals.} To compensate for inevitable errors in pre-segmentation, we introduce cross-attention to provide the missing residuals. For the sequence about the horizontal plane, we extract the potential beneficial residuals as follows:
\begin{eqnarray}
\begin{array}{cc}   
Attention(\mathcal{Q}_{h}'', \mathcal{K}_{v}'', \mathcal{V}_{v}'') = softmax(\frac{\mathcal{Q}_{h}''(\mathcal{K}_{v}'')^{T}}{\sqrt{d_{k}}}\mathcal{V}_{v}'').
\end{array}
\end{eqnarray}
For the other:
\begin{eqnarray}
\begin{array}{cc}   
Attention(\mathcal{Q}_{v}'', \mathcal{K}_{h}'', \mathcal{V}_{h}'') = softmax(\frac{\mathcal{Q}_{v}''(\mathcal{K}_{h}'')^{T}}{\sqrt{d_{k}}}\mathcal{V}_{h}'').
\end{array}
\end{eqnarray}
where $\mathcal{Q}_{h}'', \mathcal{K}_{h}'', \mathcal{V}_{h}''$ are all learned from $\mathbf{Q}_{h}''$ ($\mathcal{Q}_{v}'', \mathcal{K}_{v}'', \mathcal{V}_{v}''$ from $\mathbf{Q}_{v}''$) via a fully connected layer.
Then we add the learned residuals with the two 1D sequences to implement the reconstruction.
After that, two separate fully connected layers are employed to output the predicted horizon depth from the vertical plane sequence and the room height value from the horizontal plane one, respectively.

\subsection{Objective Function}
\label{sec3_5}
Our objective function consists of two parts: one for room layout estimation and the other for plane segmentation. For the first part, we strictly follow \cite{Jiang_2022_CVPR} and it can be denoted as $\mathcal{L}_{layout}$. For the segmentation part, we apply binary cross-entropy loss $\mathcal{L}_{segment}$ as follows:
\begin{eqnarray}
\begin{array}{cc}   
\mathcal{L}_{segment}= \frac{1}{N} \sum\limits _{i\in N}(d_{i}\log_{}{\bar{d_{i}}}+(1D_{i})\log_{}{(1-\bar{d_{i}}))}
\end{array}
\end{eqnarray}
where $\bar{d_{i}}$ is the ground truth of generated segmentation label, and $d_{i}$ is the predicted value. Ultimately, we formulate the objective function as follows:
\begin{eqnarray}
\begin{array}{cc}   
\mathcal{L}= \lambda \mathcal{L}_{segment} + \mathcal{L}_{layout}
\end{array}
\end{eqnarray}
where $\lambda$ is set to 0.75 to balance different constraints.

\section{Experiments}
\label{section4}
In this section, we validate the effectiveness of our solution and compare it with existing SoTA approaches on four popular datasets.
Concretely, we conduct experiments on a single GTX 3090 GPU, and the batch size is set to 16 for training. The proposed approach is implemented with PyTorch. We choose Adam~\cite{kingma2015adam} as the optimizer and keep the default settings. The initialized learning rate is $1\times10^{-4}$. As in previous works \cite{sun2019horizonnet,Jiang_2022_CVPR}, we adopt standard left-right flipping, panoramic horizontal rotation, luminance change, and pano stretch for data augmentation during training.

\subsection{Datasets}
Four datasets are used for our experimental validation: Stanford 2D-3D~\cite{armeni2017joint}, PanoContext~\cite{zhang2014panocontext}, MatterportLayout~\cite{zou2021manhattan} and ZInd~\cite{cruz2021zillow}. 

PanoContext~\cite{zhang2014panocontext} and Stanford 2D-3D~\cite{armeni2017joint} are two commonly used datasets for indoor panoramic room layout estimation that contain 514 and 552 cuboid room layouts, respectively. Especially,  Stanford 2D-3D~\cite{armeni2017joint} is labeled by Zou $et$ $al.$ \cite{zou2018layoutnet} and has a smaller vertical FoV than other datasets. Besides, MatterportLayout \cite{zou2021manhattan} is also annotated by Zou $et$ $al.$ \cite{zou2021manhattan}, which contains 2,295 general room layouts. The final ZInd \cite{cruz2021zillow} dataset includes cuboid, general Manhattan, non-Manhattan, and non-flat ceilings layouts, which mimic the real-world data distribution better. The splits of ZInd \cite{cruz2021zillow} consists of 24,882 (for training), 3,080 (for validation), and 3,170 (for testing) panoramas. We strictly follow the same training/validation/test splits of the four datasets as in previous works for a fair comparison. 
\subsection{Evaluation Metrics}
We employ the commonly used standard evaluation metrics for a fair comparison, including corner error (CE), pixel error (PE), intersection over the union of floor shapes (2DIoU), and 3D room layouts (3DIoU). Among them, 3DIoU yields a better reflection of the accuracy of the layout estimation in 3D space. RMSE and $\delta_{1.25}$ indicate the performance of depth estimation, e.g. the horizon-depth map, 

\begin{table}[t]
\begin{center}
 \begin{tabular}{l c c c}
  \toprule
  Method& 3DIoU(\%)& CE(\%)& PE(\%) \\
    \hline
   \multicolumn{4}{c}{Train on PanoContext + Whole Stnfd.2D3D datasets}\\
   \hline LayoutNetv2\cite{zou2021manhattan}&85.02&0.63& 1.79\\
   Dula-Netv2\cite{zou2021manhattan}&83.77&0.81& 2.43\\
    HorizonNet\cite{sun2019horizonnet}&82.63&0.74& 2.17\\  
    LGT-Net\cite{Jiang_2022_CVPR}&85.16&-& -\\
    LGT-Net [w/ Post-proc]\cite{Jiang_2022_CVPR}&84.94&0.69&2.07\\
    Ours&\textbf{85.46}&-&-\\
    Ours [w/ Post-proc]&85.00&\textbf{0.69}&2.13\\
     \hline
   \multicolumn{4}{c}{Train on Stnfd.2D3D + Whole PanoContext datasets}\\
   \hline 
   LayoutNetv2\cite{zou2021manhattan}&82.66&0.83& 2.59\\
   Dula-Netv2\cite{zou2021manhattan}&\textbf{86.60}&0.67& 2.48\\
    HorizonNet\cite{sun2019horizonnet}&82.72&0.69& 2.27\\ 
    AtlantaNet\cite{pintore2020atlantanet}&83.94&0.71&2.18\\
    LGT-Net\cite{Jiang_2022_CVPR}&85.76&-&-\\
    LGT-Net [w/ Post-proc]\cite{Jiang_2022_CVPR}&86.03&0.63&2.11\\
    Ours&85.47&-&-\\
    Ours [w/ Post-proc]&85.58&0.66&\textbf{2.10}\\
  \bottomrule
 \end{tabular}
\end{center}
 \caption{Quantitative comparison results with the current SoTA solutions evaluated
on Stanford 2D-3D and PanoContext\cite{cruz2021zillow} dataset.}
 \label{tab:cp1}
 \vspace{-0.2cm}
\end{table}
\subsection{Cuboid Room Results} 
We follow LGT-Net \cite{Jiang_2022_CVPR} to use the combined dataset scheme mentioned in Zou $et$ $al.$ \cite{zou2021manhattan} to evaluate our network on cuboid datasets. We denote the combined dataset that contains training splits of PanoContext \cite{cruz2021zillow} and whole Stanford 2D-3D datasets as ``Train on PanoContext + Whole Stnfd.2D3D datasets" in Tab. \ref{tab:cp1}. Similarly, ``Train on Stnfd.2D3D + Whole PanoContext" in Tab. \ref{tab:cp1} represents that we train our network on the combined dataset that contains training splits of Stanford 2D-3D and the whole PanoContext dataset. The scheme is commonly used in previous works \cite{zou2018layoutnet,Jiang_2022_CVPR}. We also report the results with a post-processing strategy in DuLa-Net \cite{yang2019dula} that is denoted as "Ours [w/ Post-proc]".

\noindent\textbf{Comparison results.} We exhibit the quantitative comparison results on cuboid room layouts in Tab. \ref{tab:cp1}. From the first group in Tab. \ref{tab:cp1}, we can observe that our approach outperforms all the other SoTA schemes with respect to 3DIoU. But from the second group, Dula-Net v2 \cite{zou2021manhattan} offers better 3DIoU than ours. Dula-Net v2 \cite{zou2021manhattan} employs a perspective view (i.e., cubemap) that is more effective for panoramas with small vertical FoV. However, once applied to more general room layout datasets, its performance degrades significantly (it is proved with the other two datasets in Sec. \ref{sec4_4}). On the contrary, the proposed approach shows more general performance. 
\begin{table}[t]
\begin{center}
 \begin{tabular}{p{2.5cm} p{1.2cm} p{1.2cm} p{0.7cm} p{0.7cm}}
  \toprule
  Method& 2DIoU(\%) &  3DIoU(\%)& RMSE& $\delta_{1}$ \\
    \hline
  LayoutNetv2\cite{zou2021manhattan}&78.73&75.82& 0.258&0.871\\
   Dula-Netv2\cite{zou2021manhattan}&78.82&75.05& 0.291&0.818\\
    HorizonNet\cite{sun2019horizonnet}&81.71&79.11& 0.197&0.929\\    
    AtlantaNet\cite{pintore2020atlantanet}&82.09&80.02& -&-\\
    HoHoNet\cite{sun2021hohonet}&82.32&79.88& -&-\\
    LED$^{2}$-Net\cite{wang2021led}&82.61&80.14& 0.207&0.947\\
    DMH-Net \cite{zhao20223d}&81.25&78.97& -&0.925\\
    LGT-Net\cite{Jiang_2022_CVPR}&83.52&81.11& 0.204&\textbf{0.951}\\
    Ours&\textbf{84.11}&\textbf{81.70}& \textbf{0.197}&0.950\\
  \bottomrule
 \end{tabular}
\end{center}
 \caption{Quantitative comparison results with the current SoTA solutions evaluated
on MatterportLayout\cite{zou2021manhattan} dataset.}
 \label{tab:cp2}
\end{table}
\begin{table}[t]
\begin{center}
 \begin{tabular}{m{2.4cm} m{1.2cm} m{1.2cm} m{0.7cm} m{0.7cm}}
  \toprule
  Method& 2DIoU(\%) &  3DIoU(\%)& RMSE& $\delta_{1}   $ \\
    \hline
    HorizonNet\cite{sun2019horizonnet}&90.44&88.59& 0.123&0.957\\    
    LED$^{2}$-Net\cite{wang2021led}&90.36&88.49& 0.124&0.955\\
    LGT-Net\cite{Jiang_2022_CVPR}&91.77&89.95& 0.111&0.960\\
    Ours&\textbf{91.94}&\textbf{90.13}&-&-\\
  \bottomrule
 \end{tabular}
\end{center}
 \caption{Quantitative comparison results with the current SoTA solutions evaluated
on ZInD\cite{cruz2021zillow} dataset.}
 \label{tab:cp3}
 \vspace{-0.2cm}
\end{table}
\begin{figure*}[htbp]
	\centering
	\subfigure[Qualitative comparison on MatterportLayout \cite{zou2021manhattan} dataset.]{
		\begin{minipage}[t]{\linewidth}
			\centering
			\includegraphics[width=0.97\textwidth]{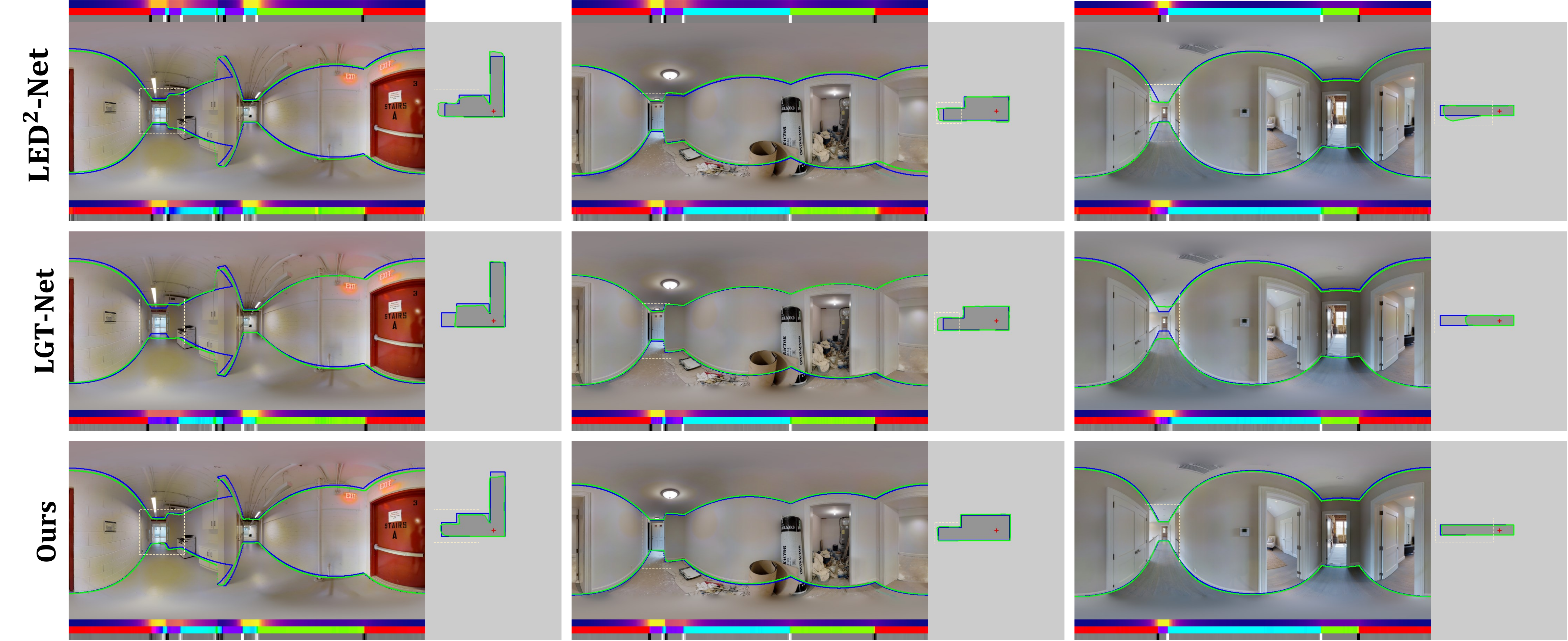}
		\end{minipage}
	}
    \subfigure[Qualitative comparison on ZInD \cite{cruz2021zillow} dataset.]{
		\begin{minipage}[t]{\linewidth}
			\centering
			\includegraphics[width=0.97\textwidth]{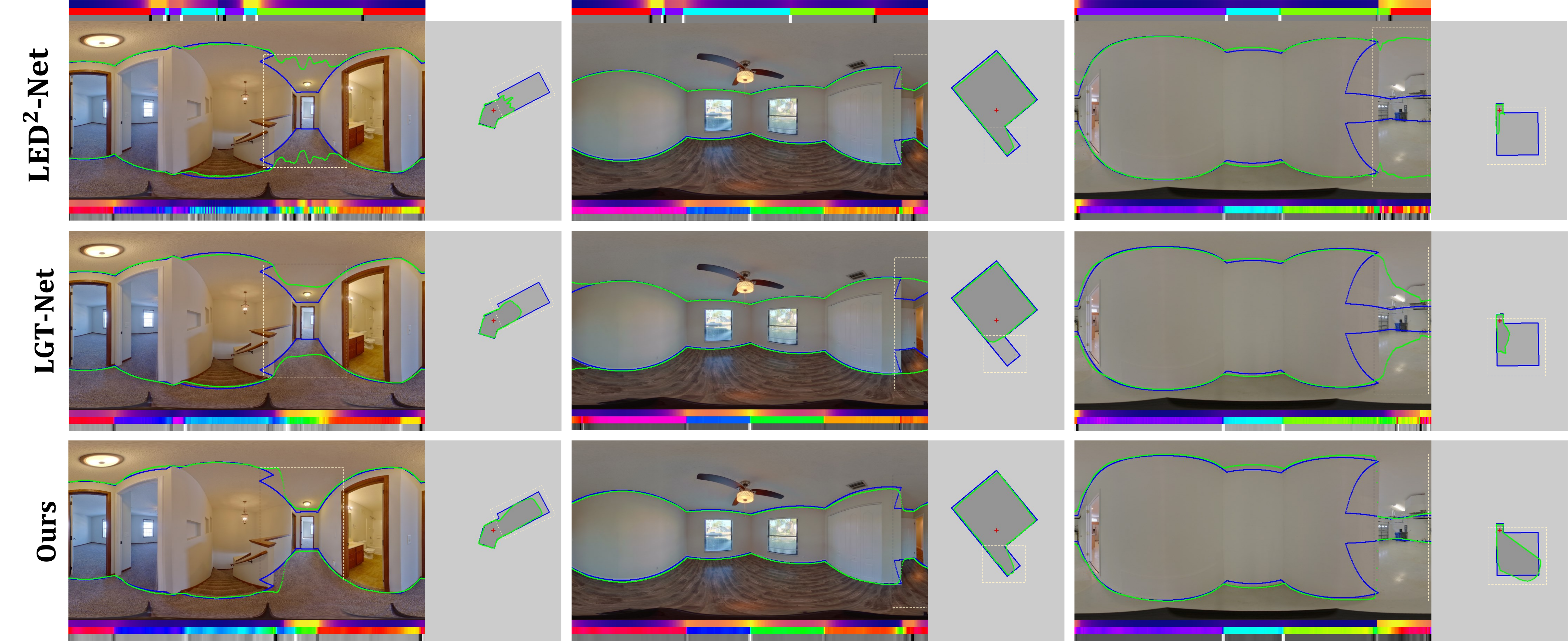}
		\end{minipage}
	}
	\caption{Qualitative comparison results evaluated on general layout datasets, MatterportLayout \cite{zou2021manhattan} and ZInD \cite{cruz2021zillow}. We compare our method with LED$^2$-Net \cite{wang2021led} and LGT-Net \cite{Jiang_2022_CVPR}. The compared methods are all not employed the post-processing strategy. The boundaries of the room layout on a panorama are shown on the left and the floor plan is on the right. Ground truth is best viewed in {\color{blue} Blue lines}, and the prediction in {\color{green} Green}. The predicted horizon depth, normal, and gradient are visualized below each panorama, and the ground truth is in the first row. We labeled the significant differences with dashed lines.}
	\label{fig:quli}
\vspace{-0.2cm}
\end{figure*}
\begin{figure*}[htbp]
\centering
\subfigure[Standford 2D-3D \cite{armeni2017joint}]{
\includegraphics[width=0.48\textwidth]{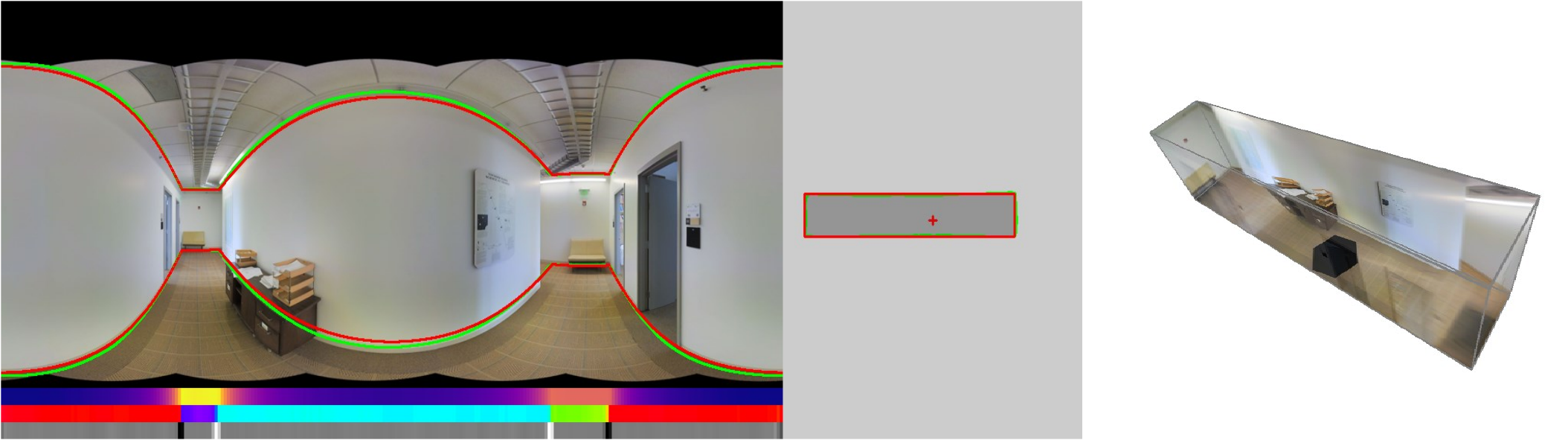}
}%
\subfigure[PanoContext \cite{zhang2014panocontext}]{
\includegraphics[width=0.48\textwidth]{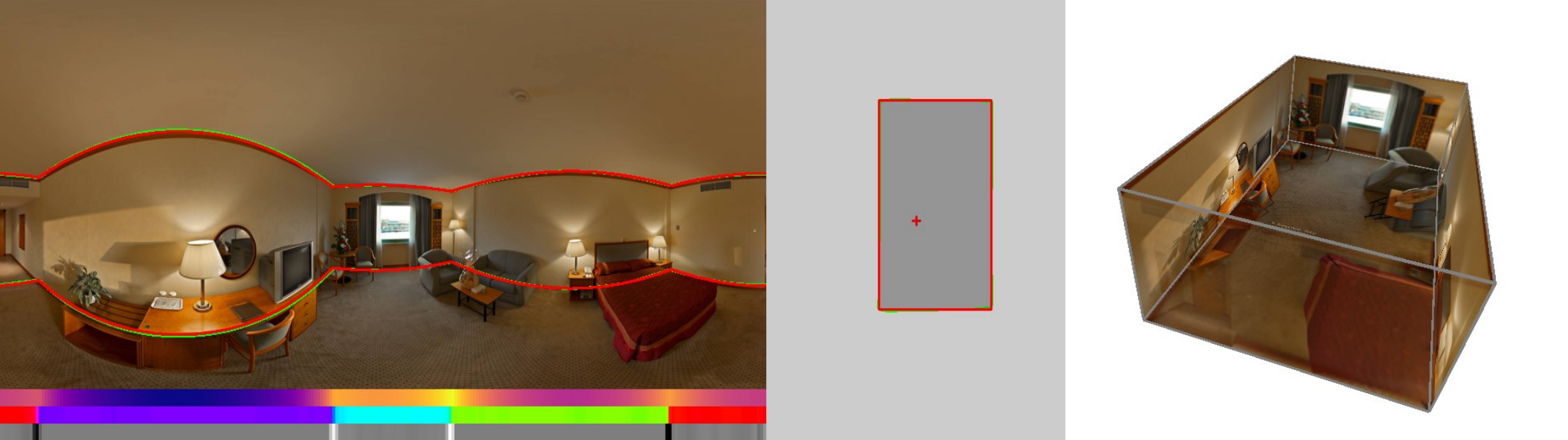}
}%

\subfigure[MatterportLayout \cite{zou2021manhattan}]{
\includegraphics[width=0.48\textwidth]{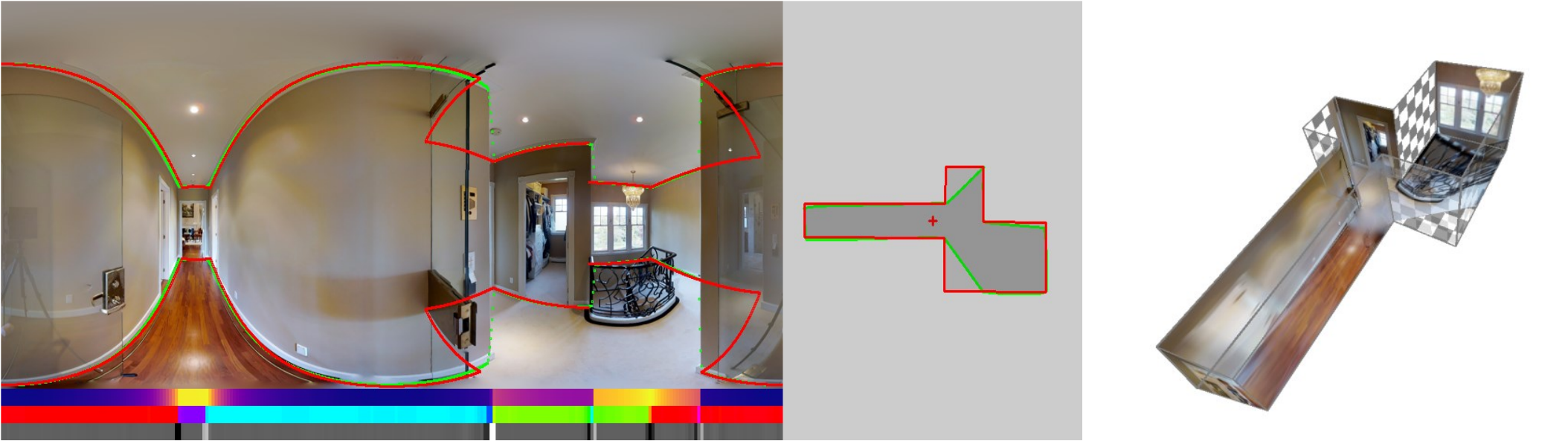}
}%
\subfigure[ZInD \cite{cruz2021zillow}]{
\includegraphics[width=0.48\textwidth]{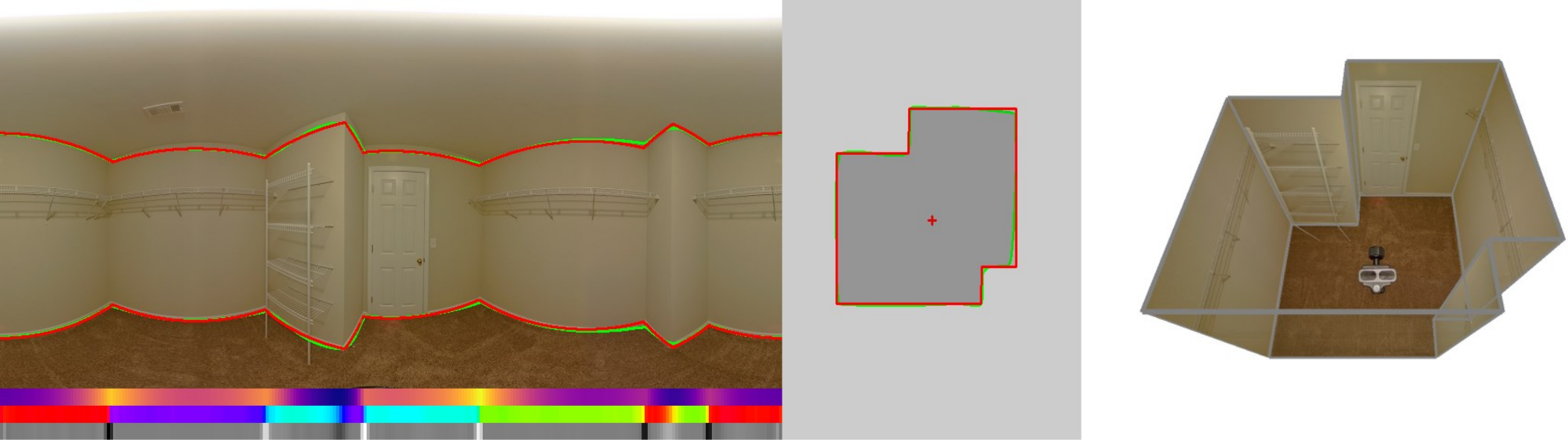}
}%
\centering
\caption{The 3D visualization results. We exhibit the predicted boundaries (best viewed in {\color{green}Green lines}) and the ones with post-processing of the prediction (best viewed in {\color{red}Red lines}) in the panoramas.}
\label{fig:3dvis}
\vspace{-0.1cm}
\end{figure*}
\begin{figure*}[t]
  \centering
  \includegraphics[width=0.97\textwidth]{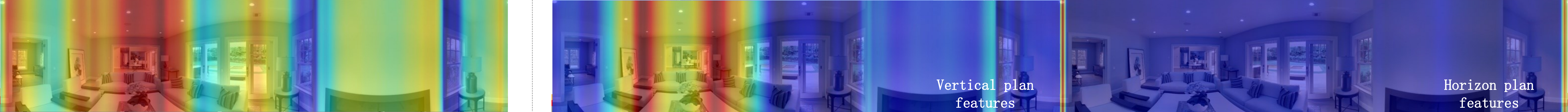} 
  \caption{We exhibit the features of the 1D representation. The one without orthogonal plane disentanglement is shown on the left, and ours is on the right. 
  Without disentangling, the left contains redundant and confusing features that are not good for layout estimation. In contrast, our disentangled vertical plane features are more discriminative, showing more attention to the layout corners.
  } 
  \label{fig:visf}
\vspace{-0.2cm}
\end{figure*}
\subsection{General Room Results}
\label{sec4_4} MatterportLayout \cite{zou2021manhattan} dataset and ZInD \cite{cruz2021zillow} dataset provide more general indoor room layouts, which is much more challenging than the cuboid room layout datasets. Tab. \ref{tab:cp2}/Tab. \ref{tab:cp3} exhibits the evaluation on MatterportLayout \cite{zou2021manhattan}/ZInD \cite{cruz2021zillow} datasets. The results of LED$^{2}$-Net \cite{wang2021led} and HorizonNet \cite{sun2019horizonnet} are from \cite{Jiang_2022_CVPR} that we strictly followed. Particularly, Jiang $et$ $al.$ utilize their official code
to re-train and re-evaluate with the standard evaluation metrics. 

\noindent\textbf{Comparison results.} We observe that Dula-Net \cite{yang2019dula} demonstrates much worse performance on the general room layout dataset (MatterportLayout), indicating that these perspective view-based methods are hard to be adapted to the more general indoor scenarios. Moreover, these 1D sequence-based methods are better than those 2D convolution-based schemes. However, they cannot produce accurate 3D layouts from a 1D sequence due to confused plane semantics, even if they introduce more powerful relationship builders (e.g., Bi-LSTM, Transformer). Hence, these methods give worse 3DIoU. Compared with them, our approach offers better performance than all other approaches with respect to 3DIoU because ours captures an explicit 3D geometric cue by disentangling orthogonal planes, which benefits recovering layouts from clear semantics (Fig. \ref{fig:visf}).

We show the qualitative comparisons in Fig. \ref{fig:quli}. From the figure, we can observe that our method offers better 3D results (floor plan). In Fig. \ref{fig:3dvis}, our results are much similar to the post-processing version. Both indicate our proposal that disentangling orthogonal planes to capture an explicit 3D geometric cue is an effective strategy.
\begin{table}[t]
\begin{center}
 \begin{tabular}{l c c}
  \toprule
  Method& 2DIoU(\%) &  3DIoU(\%)\\
    \hline
    w/o Cross-scale interaction&83.24&80.71\\
    w/o Feature assembling&82.36&80.10\\
    \hline
    w/o Disentangling planes&83.23&80.60\\
    w/o Flipping fusion&83.01&80.35\\
    \hline
    w/o Discriminative channels&82.74&80.23\\
    w/o Long-range dependencies&83.04&80.97\\
    w/o Residuals&82.89&80.45\\
    \hline
    Ours [Full]&\textbf{83.46}&\textbf{81.34}\\
  \bottomrule
 \end{tabular}
\end{center}
 \caption{Ablation study on MatterportLayout \cite{zou2021manhattan} dataset.}
 \label{tab:abla}
 \vspace{-0.2cm}
\end{table}
\subsection{Ablation study}
\label{section42}
We exhibit ablation studies in Tab. \ref{tab:abla}, where each component of our model is evaluated on MatterportLayout \cite{zou2021manhattan}. To demonstrate the effectiveness of the proposed feature assembling mechanism, we first ablate cross-scale interaction in feature assembling (denoted as "w/o Cross-scale interaction"). Then, we further remove the distortion elimination (denoted as "w/o Feature assembling"). 
To exhibit the benefit of orthogonal plane disentanglement, we put off the proposed disentangling orthogonal planes procedure (denoted as "w/o Disentangling planes") and soft-flipping fusion, respectively. 
Finally, we show the effect of each attention mechanism (denoted as "w/o Discriminative channels", "w/o Long-range dependencies", and "w/o residuals", respectively). Specially, we train each model around 500 epochs (not the best ones). Hence, the results of "Ours [Full]"
are slightly lower than those in Tab. \ref{tab:cp2}.

\noindent\textbf{Feature assembling mechanism.} 
 From Tab. ~\ref{tab:abla}, the pipeline without cross-scale interaction shows inferior performance to the full model. Further removing the distortion elimination part, the pipeline gives worse results. It is proved that both dealing with distortions and integrating cross-scale features are essential for layout estimation. 

\noindent\textbf{Disentangling orthogonal planes.} Since the geometric cues are essential for inferring 3D information from 2D images, we propose to disentangle orthogonal planes. To prove the effectiveness, we remove that stage but preserve the flipping fusion strategy (the effectiveness of this strategy is validated separately). The results in Tab. ~\ref{tab:abla} show that this stage can make the performance more competitive. Besides, the embedding of symmetry property also contributes to the promotion of the predicted results.

\noindent\textbf{Reconstructing 1D representations.} 
The dependencies of the 1D sequences have changed when disentangling the orthogonal planes. Hence, we propose to reconstruct the 1D representations. We verify the effectiveness of each attention mechanism in turn. From Tab. \ref{tab:abla}, we can observe that all three reconstruction operations (generating discriminative channels, rebuilding long-range dependencies, and providing the missing residuals) can all benefit the overall performance. Significantly, when removing the channel-wise graph, the pipeline's performance decreases significantly, demonstrating that enforcing the network concentration on discriminative channel information can capture effective information by avoiding redundancy.

\section{Conclusion}
\label{section5}
In this paper, we propose a novel panoramic indoor room layout estimation approach. Current approaches generate a 1D representation by a vertical compression operation. We argue that this strategy confuses the semantic cues of different planes. To address this issue, we propose to disentangle orthogonal planes to capture geometric cues in 3D space. Specially, we introduce a vertical flip-fusion strategy to leverage the symmetry property of indoor room layout. Besides, our experimental results demonstrate that dealing with distortion, as well as integrating shallow and deep features, can enhance the performance. Experiments demonstrate that our algorithm significantly outperforms current SoTA methods. 

\noindent\textbf{Acknowledgement}. This work was supported by  the National Natural Science Foundation of China (Nos. 62172032,  62120106009).
{\small
\normalem
\bibliographystyle{ieee_fullname}
\bibliography{egbib}
}

\end{document}